# 3D/2D Registration of Mapping Catheter Images for Arrhythmia Interventional Assistance

Pascal Fallavollita

School of Computing, Queen's University,
Kingston, Ontario, Canada
*pascal@cs.queensu.ca*

**Abstract**
Radiofrequency (RF) catheter ablation has transformed treatment for tachyarrhythmias and has become first-line therapy for some tachycardias. The precise localization of the arrhythmogenic site and the positioning of the RF catheter over that site are problematic: they can impair the efficiency of the procedure and are time consuming (several hours). Electroanatomic mapping technologies are available that enable the display of the cardiac chambers and the relative position of ablation lesions. However, these are expensive and use custom-made catheters. The proposed methodology makes use of standard catheters and inexpensive technology in order to create a 3D volume of the heart chamber affected by the arrhythmia. Further, we propose a novel method that uses *a priori* 3D information of the mapping catheter in order to estimate the 3D locations of multiple electrodes across single view C-arm images. The monoplane algorithm is tested for feasibility on computer simulations and initial canine data.

## 1. Introduction

Severe disorders of the heart rhythm that can cause sudden cardiac death or morbidity, can be treated by radio-frequency (RF) catheter ablation, which consists of inserting a catheter inside the heart, near the area from which originates the abnormal cardiac electrical activity, and delivering RF currents through the catheter tip so as to ablate this arrhythmogenic area. The precise localization of the arrhythmogenic site and positioning of the RF catheter at that site are problematic: they can impair the efficacy of the procedure and the procedure can last many hours, especially for complex arrhythmias. To shorten the duration of RF catheter ablation and increase its efficiency, commercial systems that provide a 3D color display of the cardiac electrical activation sequence during the arrhythmia have been proposed.

These systems incorporate basket electrode arrays (**Constellation**, EPT Inc.), catheters with a balloon electrode array (**Ensite 3000**, Endocardial Solutions Inc.) and catheters with magnetic position detectors (**CARTO**$^{TM}$, Biosense Webster Inc.). A complete navigation and registration framework is also available (**CartoMerge**, Biosense Webster Inc.). All these systems including purchase of system-specific catheters are costly for hospitals. The first two technologies can map the cardiac activation sequence using data recorded during a single beat whereas the CARTO$^{TM}$ system relies on data recorded point-by-point during numerous beats, which implies that the arrhythmia must remain stable during the procedure.

Other approaches have also been proposed to guide RF ablation therapy, such as the visualization of an optically tracked catheter by making use of magnetic resonance imaging (MRI) [1-2], the combination of MRI and fluoroscopy [3], ultrasound imaging of the ablation catheter [4], the combination of ultrasound and pre-operative computer tomography (CT) [5], or preoperative imaging (CT/MRI) for ablation planning [6-7]. These approaches omit incorporating the all important electrophysiological data which allows the interventionist to determine the origin of the arrhythmia. Inverse electrocardiography, an established formulation is the imaging of the activation time map on the entire surface of the heart from ECG mapping data. A few examples computed from body surface potential measurements have been incorporated to segmented MRI images by Tilg et al. [8-9] and to biplane reconstructions of the cardiac geometry by Ghanem et al. [10-11]. DeBuck et al. [12] have constructed a patient-specific 3D anatomical model from MRI and merged it with fluoroscopic images in an augmented reality environment that enables the transfer of cardiac activation times onto the model. Cristoforetti et al. [13] have developed a strategy for the spatial registration of the coarse electroanatomic map obtained by the CARTO$^{TM}$ system, and the detailed geometrical reconstruction of the left atria and pulmonary veins retrieved from CT images. Methods based on real-time integration of electroanatomic and tomographic modalities are being developed and offer an intermodal fusion based on semi-automatic registration procedures initialized by fiducial point pairing [14-15]. The above methods make use of preoperative data which do not reflect the real-time movement of the heart at the time of intervention, and introduce a logistical problem as these technologies are not present in the electrophysiological intervention room.

Recently, we have proposed a more affordable fluoroscopic navigation, emulating CARTO$^{TM}$, by obtaining local activation times from a roving catheter whose positions are computed point-by-point from biplane fluoroscopic projections. We also proposed a novel concept by superimposing the isochronal map depicting the cardiac electrical activation sequence directly over the 2D fluoroscopic image of the heart [16], a method that the CARTO$^{TM}$ technology lacks. However, biplane fluoroscopy systems are rare in the clinical setting which in turn emphasizes the importance of developing single-view 3D reconstruction algorithms.

The focus of this paper is a continuation of our previous work, but with two significant modifications. To begin with, we consider implementing a full perspective camera model instead of the parallel projection concept so as to create a more precise 3D geometry of the heart chamber affected by the arrhythmia. Second, our aim is to answer the following question: is it possible to estimate precisely the 3D locations of the mapping catheter across all C-arm image frames in a cardiac cycle using only a single view? We propose to use *a priori* 3D information of the mapping ablation catheter positions in order to achieve this. If this is possible, then the interventionist could possibly reduce procedure time accordingly as a monoplane C-arm fluoroscope is the primary imaging modality used in clinic and can provide real-time data images as well. To our knowledge this is the first work reported on estimating 3D locations using single view sequences in order to assist catheter ablation procedures. Therefore, we tag this work as a feasibility study by presenting a practical implementation and experimental analysis using computer simulations and initial results on canine data.

## 2. Methodology

### 2.1 Central Intuition

Clinical accuracy is determined by 3D reconstruction precision in the order of 2 mm or less and we don't expect to achieve this using a monoplane sequence. As showed in [16], using only a single 2D C-arm image the recovered catheter depth had an error of about 10 mm. We believe that by incorporating temporal knowledge of the catheter coordinates may help improve depth estimation results.

Initial *a priori* 3D information of the mapping catheter can be obtained using pre-operative data from CT/MRI or via two-view reconstruction by rotating the monoplane C-arm fluoroscope at two perpendicular angulations and acquiring monoplane sequences. In this paper, we focus on the latter since the principal imaging modality used to guide cardiac ablation procedures is the C-arm fluoroscope. Like most work in the cardiac field, the 3D *a priori* coordinates are reconstructed in the diastolic cardiac phase as heart motion is considered to be minimal at this instant. We then proceed to select one of the monoplane sequences acquired and estimate the depth of the mapping catheter at each image frame. The idea is to determine if or when the 3D reconstruction becomes inaccurate. In the end, what will be required is providing a 3D visual aid to the interventionist of the heart chamber and fusing the isochronal times on it in order to help them position correctly the mapping ablation catheter on the arrhythmogenic site.

### 2.2 C-arm Fluoroscopy Calibration

**Full Perspective Camera Model:** Figure 1 shows the full perspective camera model that will be used for the 3D reconstruction problem. If we define a three dimensional point $P_{world}=[X\ Y\ Z\ 1]^T$ in the world coordinate system, then its 2D projection in an image, $m=[u\ v\ 1]^T$, is achieved by constructing a projection matrix:

$$P_{mat} = \begin{bmatrix} kf & 0 & u_o \\ 0 & kf & v_o \\ 0 & 0 & 1 \end{bmatrix} \times \begin{bmatrix} r_{11} & r_{12} & r_{13} & t_x \\ r_{21} & r_{22} & r_{23} & t_y \\ r_{31} & r_{32} & r_{33} & t_z \end{bmatrix}$$

$$m = P_{mat} P_{world} \qquad (1)$$

The intrinsic matrix of size [3x3], contains the pixel coordinates of the image center, also known as the principal point ($u_o$, $v_o$), the scaling factor *k*, which defines the number of pixels per unit distance in image coordinates, and the focal length *f* of the camera (in meters). The extrinsic matrix of size [3x4] is identified by the transformation needed to align the world coordinate system to the camera coordinate system. This means that a translation vector, *t*, and a rotation matrix, *R*, need to be found in order to align the corresponding axis of the two reference frames. Lastly, image resolution (usually mm/pixel) is calculated from the imaging intensifier size divided by the actual size of the image in pixels.

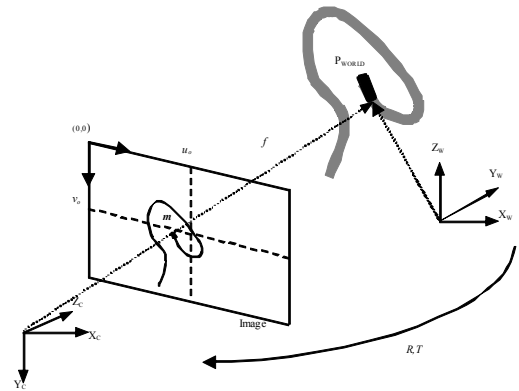

Fig 1. The perspective camera model. Any 3D world point can be projected onto a 2D plane and its coordinates would be (u, v)

pixels. The camera model is taken from the Epipolar Geometry Toolbox [17]. Image resolution would be equal to C-arm intensifier size divided by image size in pixels.

**Parallel & Weak Perspective Camera Models:** An orthographic camera is one that uses parallel projection to generate a two dimensional image of a three dimensional object. The image plane is perpendicular to the viewing direction. Parallel projections are less realistic than full perspective projections, however they have the advantage that parallel lines remain parallel in the projection, and distances are not distorted by perspective foreshortening. The parallel projection matrix is given by:

$$P_{affine} = \begin{bmatrix} k*r_{11} & k*r_{12} & k*r_{13} & (k*t_x)+u_o \\ k*r_{21} & k*r_{22} & k*r_{23} & (k*t_y)+v_o \\ 0 & 0 & 0 & 1 \end{bmatrix} \quad (2)$$

The weak perspective camera is an approximation of the full perspective camera, with individual depth points $Z_i$ replaced by an average depth $Z_{avg}$. We define the average depth, $Z_{avg}$ as being located at the centroid of the cloud of 3D points in the world coordinate system. The weak perspective projection matrix is given by:

$$P_{weak} = \begin{bmatrix} f*k*r_{11} & f*k*r_{12} & f*k*r_{13} & (f*k*t_x)+(u_o*Z_{avg}) \\ f*k*r_{21} & f*k*r_{22} & f*k*r_{23} & (f*k*t_y)+(v_o*Z_{avg}) \\ 0 & 0 & 0 & Z_{avg} \end{bmatrix}$$

$$\text{where} \quad Z_{avg} = ([r_{31}\ r_{32}\ r_{33}]^T \times centroid) + t_z \quad (3)$$

## 2.3 Mapping Catheter Segmentation

We have recently developed a four-step filter in order to enhance coronary arteries visible in C-arm fluoroscopy images [18]. Optimal filter parameters are presented there and omitted in this paper for brevity. We applied the same filter to enhance the catheters and extract electrode coordinates. Following is a brief description of these filters.

**Homomorphic Filtering:** A homomorphic filter is first used to denoise the fluoroscopic image. The illumination component of an image is generally characterized by slow spatial variation. The reflectance component of an image tends to vary abruptly. The homomorphic filter tends to decrease the contribution made by the low frequencies and amplify the contribution of high frequencies. The result is simultaneous dynamic range compression and contrast enhancement. The homomorphic filter is given by:

$$H(u,v) = (\gamma_H - \gamma_L)(1 - e^{-c \cdot (D^2(u,v)/D_o^2)}) + \gamma_L \quad (4)$$

with $\gamma_L < 1$ and $\gamma_H > 1$. The coefficient $c$ controls the sharpness of the slope at the transition between high and low frequencies, whereas $D_o$ is a constant that controls the shape of the filter and $D(u,v)$ is the distance in pixels from the origin of the filter.

**Perona-Malik Filtering:** The Perona-Malik filter is implemented here in order to reduce and remove both noise and texture from the image, as well as, to preserve and enhance structures. The diffusion equation is given by

$$\frac{\partial I}{\partial t} = div(c(x,y,t)\nabla I) \quad (5)$$

Where $I$ is the input image and $c(x, y, t)$, the diffusion coefficient, will control the degree of smoothing at each pixel point in the image. The diffusion coefficient is a monotonically decreasing function of the image gradient magnitude. It allows for locally adaptive diffusion strengths; edges are selectively smoothed or enhanced based on the evaluation of the diffusion function. Although any monotonically decreasing continuous function of the gradient would suffice as a diffusion function, two functions have been suggested:

$$c(x,y;t) = \exp(-(\frac{|\nabla I|}{K})^2) \quad (6)$$

$K$ is referred to as the diffusion constant or the flow constant. The greatest flow is produced when the image gradient magnitude is close to the value of $K$. Therefore, by choosing $K$ to correspond to gradient magnitudes produced by noise, the diffusion process can be used to reduce noise in images.

**Complex Shock Filtering:** The complex shock filter couples shock and linear diffusion in the discrete domain, showing that the process converges to a trivial constant steady state. To regularize the shock filter, the authors suggest adding a complex diffusion term and using the imaginary value as the controller for the direction of the flow instead of the second derivative. The complex shock filter is given by

$$I_t = -\frac{2}{\pi}\arctan(a\,\text{Im}(\frac{I}{\theta}))|\nabla I| + \lambda I_{\eta\eta} + \tilde{\lambda} I_{\xi\xi} \quad (7)$$

where $a$ is a parameter that controls the sharpness of the slope, $\lambda = r^{ei\theta}$ is a complex scalar, $\tilde{\lambda}$ is a real scalar, $\xi$ is the direction perpendicular to the gradient and $\eta$ is the direction of the gradient. As fluoroscopy images have a low signal to noise ratio they tend to be noisy with artifacts present in them. Hence, we believe that applying a complex filter will result in a robust and stable deblurring process for the images as the filter is effective in very noisy environments.

IJCSI

**Morphological Operation:** Morphological filtering was applied as a final image processing step in order to eliminate background elements around the object of interest. The structuring element consists of a pattern specified as the coordinates of a number of discrete points relative to a defined origin. We chose a disk structuring element that has a radius of a few pixels, since the contours of the catheters and electrodes can be modeled as a disk.

**Electrode Segmentation & Convex Hull:** Using MatLab, the 4-step filter was implemented and applied on both monoplane C-arm datasets, beginning with the diastolic image. The smoothed images were then automatically thresholded using Otsu's method. We then labeled connected components in this image using the *bwlabel* function in MatLab. The centroid of the electrodes in each labeled region was then calculated using the *regionprops* function so as to obtain the coordinates required in the two C-arm views. The algorithm outputs the centroid automatically and in case of failure, due to electrodes or catheters overlapping, we manually segment the image to obtain the desired coordinates. The convex hull algorithm is a classical and popular method used to reconstruct a 3D object from an unorganized set of points. The 2D version of this algorithm (which is easier to understand) is described as follows. Let *S* be a finite set of two-dimensional points in the plane. The convex hull of *S* is the smallest convex set that contains *S*. This means that the boundary of the convex hull is a convex polygon, whose vertices are points of *S*, and whose edges are line segments joining pairs of points of *S*.

### 2.4 *A Priori* 3D Monoplane Algorithm

We suppose that we have at our disposal a set of *n* 3D ablation catheter electrode coordinates ($X0_n$, $Y0_n$, $Z0_n$) at time $t = 0$ obtained from two-view fluoroscopic data. These *a priori* coordinates are expressed in the camera reference frame in order to have a Z-direction corresponding to catheter electrode depth. Secondly, we have at our disposal the C-arm fluoroscope gantry parameters which can be extracted from the image header DICOM files. These parameters enable us to construct a projection matrix for a specific viewing angle. By selecting one of the two acquired monoplane datasets, we can solve for the 3D displacements ($dx_{i,i+1}$, $dy_{i,i+1}$, $dz_{i,i+1}$) between consecutive C-arm image frames beginning with the first image $i=1$. Expanding equation (1) and using an additional C-arm image $i = 2$, we obtain our first two equations as follows

$$u_2 = \frac{m_1(X0+dx_{12}) + m_2(Y0+dy_{12}) + m_3(Z0+dz_{12}) + m_4}{m_9(X0+dx_{12}) + m_{10}(Y0+dy_{12}) + m_{11}(Z0+dz_{12}) + m_{12}}$$

$$v_2 = \frac{m_5(X0+dx_{12}) + m_6(Y0+dy_{12}) + m_7(Z0+dz_{12}) + m_8}{m_9(X0+dx_{12}) + m_{10}(Y0+dy_{12}) + m_{11}(Z0+dz_{12}) + m_{12}}$$

(8)

Both equations describe the pixel coordinates in the second image ($u_2$, $v_2$) and the twelve coefficients $m_i=1:12$ are the values of the projection matrix. By adding an additional C-arm image frame $i = 3$ we obtain two new equations with three additional unknowns in 3D

$$u_3 = \frac{m_1(X0+dx_{12}+dx_{23}) + m_2(Y0+dy_{12}+dy_{23}) + m_3(Z0+dz_{12}+dz_{23}) + m_4}{m_9(X0+dx_{12}+dx_{23}) + m_{10}(Y0+dy_{12}+dy_{23}) + m_{11}(Z0+dz_{12}+dz_{23}) + m_{12}}$$

$$v_3 = \frac{m_5(X0+dx_{12}+dx_{23}) + m_6(Y0+dy_{12}+dy_{23}) + m_7(Z0+dz_{12}+dz_{23}) + m_8}{m_9(X0+dx_{12}+dx_{23}) + m_{10}(Y0+dy_{12}+dy_{23}) + m_{11}(Z0+dz_{12}+dz_{23}) + m_{12}}$$

(9)

These four equations take into account the spatial positions of a projected 3D world point on the acquired C-arm images. As we have four equations with six unknowns we can extract two additional equations based on the fact that the Euclidean distances in pixels, *d*, between catheter electrode points in two consecutive images are known. It is to note that the distance between 3D points is not the same as the distance between their projected image points. Thus, we consider orthogonal projection estimations in this case and we deem that this approximation is suitable enough for the proposed analysis. We arrive at the following two equations

$$d_{12}^2 = \sqrt{(u_2-u_1)^2 + (v_2-v_1)^2}$$

$$d_{23}^2 = \sqrt{(u_3-u_2)^2 + (v_3-v_2)^2}$$

(10)

We can now solve for the three dimensional displacements. A Levenberg-Marquardt optimization scheme [19] can be used here in order to solve for the unknown displacements. For the optimization scheme, initial approximations are a requirement to initialize the process. Hence, a suitable approximation for the displacements *dx* and *dy* can be obtained if we consider a parallel back projection of the 2D image points into the world coordinate system. As for the displacements *dz*, if we assume that the average depth of the catheter electrodes remains relatively constant in consecutive time frames (i.e. weak perspective camera model), then we can calculate the average depth of the 3D points (*X0*, *Y0*, *Z0*). This average depth should be relatively the same at future time instants $t = 2, 3$, etc., signifying that depths *dz* will be equal to zero for the optimization scheme. However, for the sake of a more exhaustive analysis, we also consider depth displacements $dz \in [1\text{-}5]$ millimeters as well. To justify the range of approximations, we note that the acquisition frame rate of the fluoroscopy images was such that the movement between catheter electrodes in two adjacent images resulted in displacements less than ten pixels. Finally, we note that the proposed equations do not incorporate nonrigid constraints implying that the monoplane 3D reconstruction of rigid moving objects (i.e. catheter electrodes) might be feasible.

2.5 Clinical Data Acquisition

A mongrel dog was anesthetized and laid on its right side on a fluoroscopy table (Integris Allura, Philips Inc.). A reference catheter and a pacing catheter were inserted into the right ventricle, close to the septal wall. The role of the reference catheter was to define an origin for our 3D coordinate system. This was important as motion artifacts are ever present during the experiments (heart beat, respiration, etc.), hence we deemed it appropriate to position it near a rigid landmark so that it experiences less movement due to artifacts. The role of the pacing catheter was to produce a simple electrical activation sequence so as to validate the isochronal maps. Finally, a standard RF ablation catheter was inserted from the femoral vein into the left ventricle (LV) of the dog. During the course of the experiment, this mapping catheter was moved to 20 different sites (i.e. point-by-point as the CARTO$^{TM}$ system) within the ventricle in order to obtain electrical and geometrical data from sufficient sites to map the activation sequence. The 20 landmarks were selected to reflect as closely as possible the entire volume of the ventricle. Electrograms were recorded using the CardioMap software system (*Research Center at Sacré Coeur Hospital*, Montreal, Canada). Specifications for the acquisition software were: 1000 samples/second, 0.05 Hz high-pass and 450 Hz low-pass frequencies. Local activation time was measured (ms) as the difference in the times of the fastest negative deflections (dV/dt) seen in the two electrograms recorded with the reference catheter and the mapping catheter. The fluoroscopic image acquisition rate was set to 60 fps so as to minimize motion artifacts as interframe 2D images are thus closer to each other and yield a smaller displacement between the objects in consecutive images. Images were recorded during approximately 2 seconds at the end of the expiration. The monoplane fluoroscopic C-arm was rotated by 90˚ to acquire two images for each mapping site: a left lateral view (the C-arm in a vertical position) and a posterior view (the C-arm in a horizontal position). Images were recorded with a 512 x 512 pixel resolution. We selected the diastolic frame for each mapping site so as to minimize motion blur by identifying the frame having the smallest root-mean-square difference with the preceding frame.

2.6 Evaluation

**Computer Simulations:** In order to validate out method, we first tested our proposed procedure on synthetic experimentation. We created a 3D helix containing 30 coordinate points so as to model the shape of a catheter. Then we created two C-arm gantry setups that represented the posterior/anterior (PA) and left lateral (LAT) views of the heart. The focal length of the fluoroscopic X-ray system was equal to 1 meter and the helix location was set to 50 cm along the focal axis. The primary angles were equal to (90˚, 0˚) respectively for the PA and LAT views, whereas the secondary angles were equal to (0˚, 0˚) for both views respectively. The image sizes were set to [512x512] pixels and the intensifier size was chosen to be [178x178] millimeters. This allowed for a resolution 0.347mm/pixel. We could now calculate two projection matrices and we projected the 3D helix points on a first set of biplane images. These two-view images represent a first time instant at $t$=0. Subsequent biplane images are calculated by applying rigid movement on the 3D helix coordinates using the following rigid motion equation:

$$X_{t=2} = R_{\theta\phi\psi} \times X_{t=1} + [T_x, T_y, T_z] \quad (11)$$

The 3D angles were set to $[\theta, \Phi, \varphi] \in [0.5˚, 0.5˚, 0.5˚]$ and the 3D translations were set to $[Tx, Ty, Tz] \in [1, 1, 1]$ millimeters each. This accounted for 2D interframe displacements in the images of (-6.82, -6.03) pixels in the left lateral view and (4.97, 7.12) pixels for the posterior/anterior views. These values are similar to what was perceived using the 60 fps acquisition rate in the clinical images. A new set of biplane images and 3D helix points are obtained at $t$=1. In a similar fashion, equation (11) is reapplied to produce subsequent biplane data. Excluding the *a priori* biplane set at $t$=0, a total of five biplane datasets are generated. For good measure we added errors of up to 2 mm to the coordinates.

**Clinical Experiment:** Regarding the clinical data, as we had a total of 20 biplane datasets, we calculated the 3D catheter points using a conventional biplane reconstruction algorithm (triangulation method). Details of the biplane algorithm are omitted here as it can be deduced that by using the two projection matrices for the X-ray gantry settings and the 2D electrode coordinates, we can easily estimate the 3D world points by solving a set of 4 projection equations with 3 unknowns (X, Y, Z) when considering a single point. A complete description is available in [20]. For a specific dataset, we first extracted the diastolic images and used our 4-step filter to extract the electrodes. Then we performed two-view reconstruction. This 3D *a priori* model was then used to initialize our monoplane equations. Since we are supposing that the C-arm fluoroscope gantry parameters remain fixed in a monoplane analysis, we can then calculate projection matrices that will allow the 3D points to be projected in the images temporally. We optimized the equations using a total of 12 visible electrodes including the actual mapping tip-electrode; the one that is in contact with the ventricle wall and directs heat to the arrhythmogenic site. With respect to the initial 3D displacements for each electrode, we used the parallel projection approximation for (*dx, dy*) and depth estimates were set in the interval $dz \in$ [0-5] mm.

## 3. Results & Discussion

### 3.1 Computer simulation results

Table 1 shows the simulation results for rigid motion analysis on a helix. We observe that as the number of C-arm images used to optimize our monoplane equations increases, the overall reconstruction results deteriorate. This is expected as the uncertainty of landmark positions increases temporally in a single view framework. The initial depth approximation *dz* plays a role in the convergence process. If we select an initial depth approximation of *dz* = 1 mm, which is equal to the true simulated displacements from our computer simulations, then we obtain lower 3D root mean square errors (RMS) between optimized and true three dimensional coordinates. The RMS values are accumulated values across the number of images used for the monoplane algorithm. Depth initial estimates $dz \in$ [1,2,3] mm produce RMS values less than 3 mm when using three consecutive monoplane images. The PA view simulations produce better results probably due to the helix points being projected with no coplanarity in the 2D images. As expected, if the initial guess is unrealistic (i.e. $dz \in$ [4, 5] mm) reconstruction results deteriorate as RMS errors are larger than 10mm. Interestingly, the temporal 3D RMS values for all 30 helix points are 8.1mm an 7.2 mm for the LAT and PA views respectively when using six consecutive images and an initial guess of *dz* =3 mm. This is less than the 10 mm estimated electrode depth using only a single image in [16].

### 3.2 Clinical results

Image contrast in the clinical experiment affects the accuracy of the segmentation algorithm and depth estimates: visual analysis revealed that the LAT images always had a better contrast than the PA images because of the oblong section of the canine torso. First, we were able to automatically segment the reference and mapping catheter electrodes in 16 of 20 datasets using the proposed 4-step filter and thresholding algorithm. The remaining four images had to be manually segmented. Regarding the PA images, the filtering and thresholding algorithm failed in the majority of images and therefore we relied on a manual segmentation of the electrodes. Figure 2 shows an example of an C-arm image segmentation using the proposed filter on a LAT and PA image. Once the electrodes were extracted from the twenty datasets, we selected the mapping catheter tip electrode coordinates from the diastolic images and performed two-view triangulation and constructed a convex hull as shown in Figure 3. This 3D representation of the left ventricle combined with the electrophysiological data was obtained using a full perspective camera model.

Table 1. Computer simulation results for various depths *dz* (in mm).

| | Left Lateral View | | | | Antero-Posterior View | | | |
|---|---|---|---|---|---|---|---|---|
| $dz=0$ # Images | mean | min | max | 3D RMS | mean | min | max | 3D RMS |
| 3 | 2.188 | 0.148 | 3.840 | 3.976 | 1.033 | 0.096 | 1.917 | 1.911 |
| 4 | 1.752 | 0.094 | 3.837 | 4.527 | 0.976 | 0.003 | 2.067 | 2.613 |
| 5 | 1.463 | 0.003 | 3.951 | 5.152 | 0.929 | 0.007 | 2.128 | 3.204 |
| 6 | 1.150 | 0.015 | 3.373 | 5.324 | 0.869 | 0.004 | 2.160 | 3.674 |
| $dz=1$ # Images | mean | min | max | 3D RMS | mean | min | max | 3D RMS |
| 3 | 1.491 | 0.043 | 2.877 | 2.763 | 0.578 | 0.005 | 0.929 | 0.983 |
| 4 | 1.272 | 0.039 | 2.876 | 3.122 | 0.651 | 0.006 | 1.080 | 1.541 |
| 5 | 1.247 | 0.101 | 2.998 | 3.802 | 0.657 | 0.010 | 1.138 | 2.004 |
| 6 | 1.230 | 0.011 | 2.413 | 4.411 | 0.641 | 0.002 | 1.165 | 2.398 |
| $dz=2$ # Images | mean | min | max | 3D RMS | mean | min | max | 3D RMS |
| 3 | 1.254 | 0.005 | 1.913 | 2.048 | 0.915 | 0.060 | 1.830 | 1.742 |
| 4 | 1.223 | 0.098 | 2.628 | 2.813 | 0.977 | 0.002 | 1.996 | 2.592 |
| 5 | 1.363 | 0.006 | 3.052 | 4.095 | 1.045 | 0.003 | 2.070 | 3.443 |
| 6 | 1.522 | 0.021 | 3.327 | 5.645 | 1.114 | 0.001 | 2.109 | 4.314 |
| $dz=3$ # Images | mean | min | max | 3D RMS | mean | min | max | 3D RMS |
| 3 | 1.190 | 0.059 | 2.860 | 2.341 | 1.890 | 1.050 | 2.794 | 3.142 |
| 4 | 1.418 | 0.008 | 3.643 | 3.873 | 1.942 | 0.896 | 2.954 | 4.458 |
| 5 | 1.730 | 0.009 | 4.072 | 5.787 | 2.004 | 0.842 | 3.023 | 5.815 |
| 6 | 2.074 | 0.012 | 4.351 | 8.097 | 2.074 | 0.825 | 3.057 | 7.220 |
| $dz=4$ # Images | mean | min | max | 3D RMS | mean | min | max | 3D RMS |
| 3 | 1.698 | 0.014 | 3.862 | 3.390 | 2.866 | 2.040 | 3.758 | 4.635 |
| 4 | 2.197 | 0.007 | 4.659 | 5.567 | 2.913 | 1.885 | 3.912 | 6.471 |
| 5 | 2.568 | 0.064 | 5.091 | 8.037 | 2.974 | 1.832 | 3.976 | 8.363 |
| 6 | 3.021 | 0.465 | 5.375 | 10.979 | 3.042 | 1.820 | 4.006 | 10.314 |
| $dz=5$ # Images | mean | min | max | 3D RMS | mean | min | max | 3D RMS |
| 3 | 2.668 | 0.978 | 4.862 | 4.714 | 3.842 | 3.030 | 4.723 | 6.154 |
| 4 | 3.173 | 0.969 | 5.673 | 7.476 | 3.885 | 2.873 | 4.872 | 8.528 |
| 5 | 3.544 | 0.812 | 6.111 | 10.492 | 3.944 | 2.822 | 4.930 | 10.965 |
| 6 | 4.004 | 1.424 | 6.399 | 14.028 | 4.011 | 2.814 | 4.955 | 13.468 |

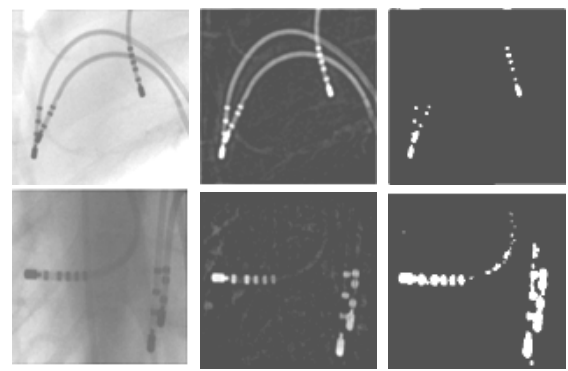

Fig 2. Example of segmentation algorithm using (*top row*) left lateral view and (*bottom row*) the posterior anterior view. The reference and pacing catheters are adjacent to each other in the right ventricle; the third catheter is the ablation catheter in the left ventricle. (*Left column*): original cropped image; (*Center column*): the electrodes are enhanced and background suppressed using a 4-step filter; (*Right column*): segmentation of the filtered image using Otsu's method and labeling. Manual segmentation would be necessary for the PA image.

Table 2. Various depth estimations of catheter electrodes using 3 and 6 consecutive X-ray images over the 20 datasets. ( in mm).

| | | Left Lateral View | | | | | | | | Posterior-Antero View | | | | | | | |
|---|---|---|---|---|---|---|---|---|---|---|---|---|---|---|---|---|---|
| | | Mapping electrode | | | | Twelve electrodes | | | | Mapping electrode | | | | Twelve electrodes | | | |
| *dz=0* | | | | | | | | | | | | | | | | | |
| | | mean | min | max | 3D RMS | mean | min | max | 3D RMS | mean | min | max | 3D RMS | mean | min | max | 3D RMS |
| # Images | 3 | 0.498 | 0.295 | 0.725 | 0.791 | 0.470 | 0.131 | 0.887 | 0.755 | 0.977 | 0.654 | 1.300 | 1.435 | 0.931 | 0.322 | 1.674 | 1.418 |
| | 6 | 0.533 | 0.121 | 1.075 | 1.397 | 0.553 | 0.044 | 1.311 | 1.449 | 0.958 | 0.249 | 1.814 | 2.642 | 0.994 | 0.055 | 2.295 | 2.762 |
| *dz=1* | | | | | | | | | | | | | | | | | |
| | | mean | min | max | 3D RMS | mean | min | max | 3D RMS | mean | min | max | 3D RMS | mean | min | max | 3D RMS |
| # Images | 3 | 1.021 | 0.797 | 1.240 | 1.615 | 1.031 | 0.628 | 1.411 | 1.624 | 1.596 | 1.157 | 2.020 | 2.567 | 1.580 | 0.815 | 2.450 | 2.566 |
| | 6 | 1.039 | 0.494 | 1.721 | 3.554 | 1.050 | 0.332 | 1.948 | 3.609 | 1.578 | 0.634 | 2.585 | 5.190 | 1.626 | 0.386 | 3.060 | 5.323 |
| *dz=2* | | | | | | | | | | | | | | | | | |
| | | mean | min | max | 3D RMS | mean | min | max | 3D RMS | mean | min | max | 3D RMS | mean | min | max | 3D RMS |
| # Images | 3 | 1.902 | 1.641 | 2.164 | 3.042 | 1.942 | 1.485 | 2.360 | 3.095 | 2.581 | 2.105 | 3.056 | 4.138 | 2.560 | 1.763 | 3.436 | 4.109 |
| | 6 | 1.968 | 1.331 | 2.683 | 6.730 | 1.993 | 1.143 | 2.935 | 6.846 | 2.509 | 1.522 | 3.557 | 8.399 | 2.467 | 1.193 | 3.855 | 8.355 |
| *dz=3* | | | | | | | | | | | | | | | | | |
| | | mean | min | max | 3D RMS | mean | min | max | 3D RMS | mean | min | max | 3D RMS | mean | min | max | 3D RMS |
| # Images | 3 | 2.876 | 2.616 | 3.136 | 4.578 | 2.918 | 2.460 | 3.334 | 4.635 | 3.566 | 3.083 | 4.040 | 5.693 | 3.531 | 2.731 | 4.403 | 5.638 |
| | 6 | 2.933 | 2.297 | 3.645 | 9.922 | 2.957 | 2.113 | 3.898 | 10.029 | 3.461 | 2.394 | 4.533 | 11.605 | 3.501 | 2.040 | 5.000 | 11.718 |
| *dz=4* | | | | | | | | | | | | | | | | | |
| | | mean | min | max | 3D RMS | mean | min | max | 3D RMS | mean | min | max | 3D RMS | mean | min | max | 3D RMS |
| # Images | 3 | 3.848 | 3.592 | 4.103 | 6.113 | 3.893 | 3.434 | 4.310 | 6.177 | 4.550 | 4.067 | 5.033 | 7.250 | 4.528 | 3.734 | 5.387 | 7.215 |
| | 6 | 3.912 | 3.269 | 4.630 | 13.173 | 3.923 | 3.082 | 4.859 | 13.201 | 4.432 | 3.359 | 5.507 | 14.824 | 4.469 | 2.988 | 5.970 | 14.932 |
| *dz=5* | | | | | | | | | | | | | | | | | |
| | | mean | min | max | 3D RMS | mean | min | max | 3D RMS | mean | min | max | 3D RMS | mean | min | max | 3D RMS |
| # Images | 3 | 4.795 | 4.534 | 5.057 | 7.609 | 4.871 | 4.406 | 5.291 | 7.724 | 5.536 | 5.049 | 6.022 | 8.808 | 5.519 | 4.710 | 6.376 | 8.767 |
| | 6 | 4.885 | 4.242 | 5.597 | 16.399 | 4.899 | 4.063 | 5.827 | 16.468 | 5.405 | 4.330 | 6.482 | 18.051 | 5.440 | 3.959 | 6.939 | 18.149 |

Table 2 presents the values for the 3D monoplane analysis using three and six consecutive images. Once again we validated our algorithm using various initial depth estimates $dz \in [0-5]$ mm. We performed two-view reconstruction at each image frame beginning with the diastolic image allowing us to have ground truth for the true 3D displacements and 3D reconstruction between consecutive images. We also determined that the minimum average 3D displacement between consecutive images for the twenty datasets was 0.10 mm and that the maximum average 3D displacement of the catheter electrodes was equal to 1.63 mm. This useful information allows us to hypothesize that the best results for our monoplane analysis should come from initial depth approximations in the range of $dz \in [1, 2]$ mm.

From Table 2 we observe that the LAT views used for monoplane reconstruction provided better recovered displacements and lower 3D accumulated RMS values when compared to the PA view. The 3D RMS values increased as we doubled the number of consecutive images used in the algorithm. The algorithm produced 3D RMS errors lower than 4.1 mm using $dz \in [0, 1, 2]$ mm initial approximations when considering both the LAT and PA views and three successive images. When considering six consecutive images, a maximum RMS error equal to 8.36 mm was obtained. Depending on the initial guess used to solve our equations our results are up to five times better than in [16] where we obtained 15 mm (posterior/anterior view) and 10 mm (left lateral view) depth errors for the ablation catheter tip electrode using a single C-arm image. Even when using six consecutive images and an initial depth error of $dz = 3$ mm we obtain 3D accumulated RMS errors of about 10 mm in LAT and 12 mm in PA views respectively which is again more accurate than [16]. Nevertheless, our method relies on initial *a priori* 3D information of the catheter, this implies more preprocessing time. Also, we could not attain our goal of estimating accurately the 3D coordinates of the electrodes across all C-arm images in a cardiac cycle. Depending on the acquisition rate, the total number of images making up a cardiac cycle can be at most 15. We thus conclude that we can recover adequately the rigid movement of the electrodes up to six consecutive images and that nonrigid equations should be incorporated in our proposed monoplane algorithm.

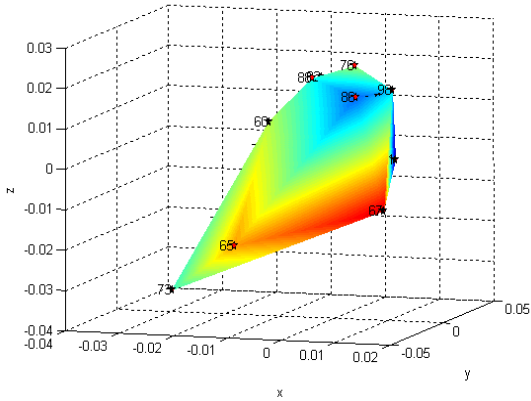

Fig 3. Three dimensional depiction of the left ventricle using a full perspective model and convex hull. The local activation times measured with the mapping catheter and colored regions having the same local activation times are superimposed on the model. The star (*) symbol represents the tip electrode of the mapping catheter which was moved to 20 different locations.

The clinical strategy is to provide the electrophysiologist with a real time representation of the activation times during catheter ablation procedures, as well as the 3D volume of the heart chamber obtained by a mapping catheter. As an addition to the CARTO$^{TM}$ technology, we register 3D isochronal information directly on the 2D C-arm images. Figure 4 and Figure 5 show the fusion of the 3D activation model with the fluoroscopic images. For both figures, the image depicts the mapping catheter tip electrode with a local activation time of 73 milliseconds. Figure 4 shows two local activation times that are inaccurate (56 ms and 61 ms); similarly in Figure 5, we observe two local activation times (79 ms and 94 ms) which are no longer part of the convex hull before projection. The smallest activation times are positioned

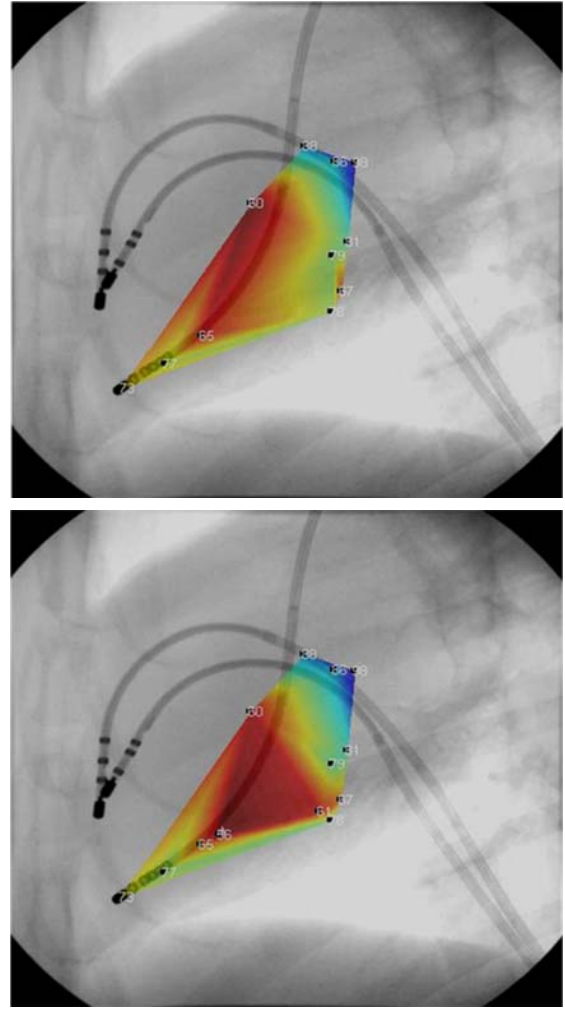

Fig 4. Fusion of electrophysiology on left lateral images of a mongrel dog. (*Top*) Visible electrophysiological data on exact depth coordinates of the ablation catheter. (*Bottom*) Erroneous electrophysiological data on estimated depth coordinates using our monoplane three view analysis.

closer to the pacing catheter in the left ventricle, as would be expected.

Even though monoplane reconstruction yielded depth errors, the isochronal maps were close enough to ground truth which raises promise for our monoplane technique. We are optimistic that in the near future this method can be validated on actual patient data. When compared to [1-13], our method uses readily available intraoperative information in order to calculate the 3D coordinates of the mapping catheters in real time. We use this

IJCSI

information to estimate catheter electrode positions in a single view sequence and determine the feasibility of monoplane 3D reconstruction as adequate when analyzing rigid moving objects. When compared to the CARTO™ system we are cost effective by making use of standard catheters and existing technology (i.e. monoplane C-arm fluoroscope), we attempt to detect more than one electrode across monoplane C-arm sequences, and we fuse 2D isochronal maps directly in the fluoroscopy images.

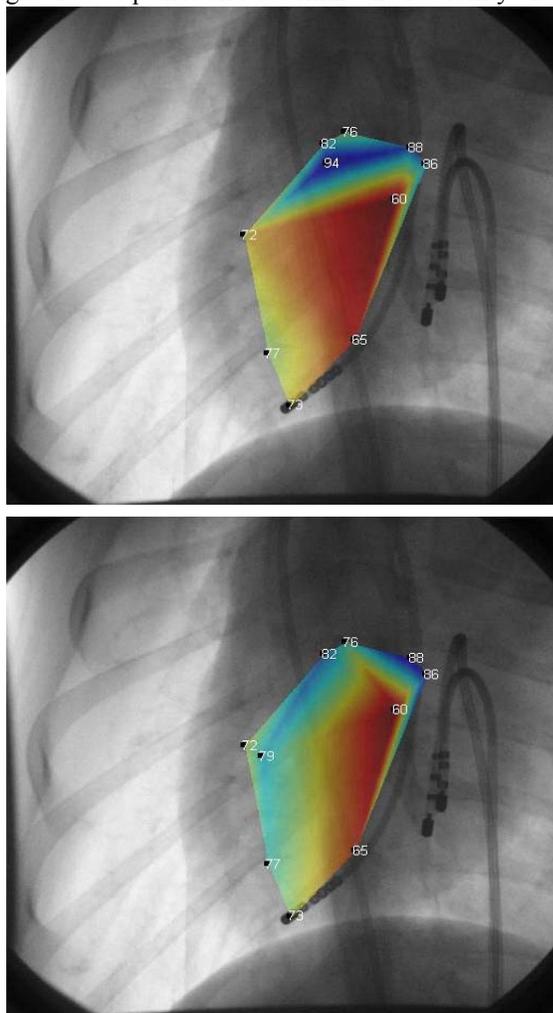

Fig 5. Fusion of electrophysiology on posterior/anterior images of a mongrel dog. (*Top*) Visible electrophysiological data on exact depth coordinates of the ablation catheter. (*Bottom*) Erroneous electrophysiological data on estimated depth coordinates using our monoplane three view analysis.

## 4. Conclusion

A novel method to estimate the depth of the mapping catheter was presented. By assuming that *a priori* 3D model is present at a first time instant, we can estimate the depth of the catheter by making use of consecutive single view fluoroscopy images. Our feasibility study provided results that were an improvement of up to 50% when compared to the two-view projection method developed in [16] which use only a single image. Activation times were fused directly on the C-arm images in order to illustrate the evolution of cardiac electrical propagation. By exploiting spatial and projective information using only single plane sequences, we aim to decrease overall intervention time and still maintain high level accuracy when predicting the depth position of the mapping catheter. Future work will focus on adding additional nonrigid constraints to our monoplane equation which capture the inherent movement of the heart. With some additional work, we are confident that our proposed analysis can be adapted to meet the demands of MRI/C-arm or CT/C-arm registration in providing a navigational aid when treating any sustained arrhythmia.

### Acknowledgments

The author would like to thank Dr. Pierre Savard, from École Polytechnique, Canada, for providing the image datasets.

### References

[1] Lardo AC, Halperin H, Yeung C, Jumrussirikul P, Atalar E, McVeigh E (1998) Magnetic resonance guided radiofrequency ablation: Creation and visualization of cardiac lesions. Med Image Comput Comput Assist Interv Int Conf 1496:189-196

[2] Lardo A, McVeigh E, Jumrussirikul P et al (2000) Visualization and temporal/spatial characterization of cardiac radiofrequency ablation lesions using magnetic resonance imaging. Circulation 102(6): 698-705

[3] Chu E, Fitzpatrick A, Chin M et al (1994) Radiofrequency catheter ablation guided by intracardiac echocardiography. Circulation 89(3):1301-1305

[4] Sun Y, Kadoury S, Li Y et al (2007) Image guidance of intracardiac ultrasound with fusion of pre-operative images. MICCAI 1: 60-67

[5] Razavi R, Hill DLG, Keevil SF et al (2003) Cardiac catheterization guided by MRI in children and adults with congenital heart disease. Lancet 362(9399): 1877-1882

[6] Kato R (2003) Pulmonary vein anatomy in patients undergoing catheter ablation of atrial fibrillation: lessons learned by use of magnetic resonance imaging. Circulation 107(15): 2004-2010

[7] Lacomis JM, Wigginton W, Fuhrman C et al (2003) Multi-detector row CT of the left atrium and pulmonary veins before radio-frequency catheter ablation for atrial fibrillation. Radiographics 23: S35-S48

[8] Tilg B, Fischer G, Modre R et al (2003) Electrocardiographic imaging of atrial and ventricular electrical activation. Med Image Anal 7(3): 391-398

[9] Modre R, Tilg B, Fischer G, Wach P (2001) An iterative algorithm for myocardial activation time imaging. Comput Methods Programs Biomed 64(1): 1-7

[10] Ghanem R, Jia P, Rudy Y (2003) Heart-surface reconstruction and ECG electrodes localization using fluoroscopy, epipolar geometry and stereovision: Application to noninvasive imaging of cardiac electrical activity. IEEE Trans Med Imag 22(10):1307-1318

[11] Ramanathan C, Ghanem R, Jia P, Rudy K (2004) Noninvasive electrocardiographic imaging for cardiac electrophysiology and arrhythmia. Nature Med 10(4): 422-428

[12] De Buck S, Maes F, Ector J et al (2005) An augmented reality system for patient-specific guidance of cardiac catheter ablation procedures. IEEE Trans Med Imaging 24:1512-1524

[13] Cristoforetti A, Masè M, Faes L et al (2007) A stochastic approach for automatic registration and fusion of left atrial electroanatomic maps with 3D CT anatomical images. Phys Med Biol. 52(20):6323-6337

[14] Tops LF, Bax JJ, Zeppenfeld K et al (2005) Fusion of multislice computed tomography imaging with three-dimensional electroanatomic mapping to guide radiofrequency catheter ablation procedures. Heart Rhythm 2: 1076-81

[15] Sra J (2005) Registration of three dimensional left atrial images with interventional systems. Heart 91:1098-1104

[16] Fallavollita P, Savard P, Sierra G (2004) Fluoroscopic navigation to guide RF catheter ablation of cardiac arrhythmias. IEEE 26th Annual International Conference of the Engineering in Medicine and Biology Society 1:1929-1932

[17] Mariottini GL, Prattichizzo D (2005) EGT for Multiple View Geometry and Visual Servoing. Robotics and Vision with Pinhole and Panoramic Cameras. IEEE Robotics and Automation Magazine 12(4):26-39

[18] Fallavollita P, Cheriet F (2006) Towards an automatic coronary artery segmentation algorithm.  IEEE 28th Annual International Conference of the Engineering in Medicine and Biology Society 3037-3040

[19] Marquardt DW (1963) An algorithm for least-squares estimation of nonlinear parameters. SIAM J. Appl. Math 11(2):431-441

[20] Fallavollita P, Cheriet F (2008) Optimal 3D Reconstruction of Coronary Arteries for 3D Clinical Assessment. Comput Med Imaging Graph, 32(6):476-487

**Pascal Fallavollita** received his B.Eng. degree in Mechanical Engineering at McGill University in 2002 and Ph.D degree in Biomedical Engineering at École Polytechnique de Montréal in 2008. He is currently a postdoctoral fellow at the School of Computing at Queen's University, Canada. His main research interests lie in the areas of medical image processing and computer aided surgery. His clinical focus lies in cardiac arrhythmia ablation, coronary angiography and prostate brachytherapy procedures.